\documentclass[journal]{IEEEtran}
\makeatletter\if@twocolumn\PassOptionsToPackage{switch}{lineno}\else\fi\makeatother

\usepackage{graphicx}
\usepackage{xurl}
\usepackage[T1]{fontenc}
\ifCLASSINFOpdf
\else
\fi
\usepackage{amsmath}
\usepackage[cmintegrals]{newtxmath}
\usepackage{url,multirow,morefloats,floatflt,cancel,tfrupee}
\makeatletter

\AtBeginDocument{\@ifpackageloaded{textcomp}{}{\usepackage{textcomp}}}
\makeatother
\usepackage{colortbl}
\usepackage{xcolor}
\usepackage{pifont}
\usepackage[nointegrals]{wasysym}
\makeatletter

\def\mcWidth#1{\csname TY@F#1\endcsname+\tabcolsep}

\def\cAlignHack{\rightskip\@flushglue\leftskip\@flushglue\parindent\z@\parfillskip\z@skip}
\def\rAlignHack{\rightskip\z@skip\leftskip\@flushglue \parindent\z@\parfillskip\z@skip}

\@ifundefined{etal}{}{}

\usepackage{ifxetex}
\ifxetex\else\if@twocolumn\@ifpackageloaded{stfloats}{}{\usepackage{dblfloatfix}}\fi\fi

\AtBeginDocument{
\expandafter\ifx\csname eqalign\endcsname\relax
\def\eqalign#1{\null\vcenter{\def\\{\cr}\openup\jot\m@th
  \ialign{\strut$\displaystyle{##}$\hfil&$\displaystyle{{}##}$\hfil
      \crcr#1\crcr}}\,}
\fi
}

\AtBeginDocument{%
  \@ifpackageloaded{endfloat}%
   {\renewcommand\efloat@iwrite[1]{\immediate\expandafter\protected@write\csname efloat@post#1\endcsname{}}}{\newif\ifefloat@tables}%
}%

\def\BreakURLText#1{\@tfor\brk@tempa:=#1\do{\brk@tempa\hskip0pt}}
\let\lt=<
\let\gt=>
\def\processVert{\ifmmode|\else\textbar\fi}

\@ifundefined{subparagraph}{
\def\subparagraph{\@startsection{paragraph}{5}{2\parindent}{0ex plus 0.1ex minus 0.1ex}%
{0ex}{\normalfont\small\itshape}}%
}{}

\newcommand\role[1]{\unskip}
\newcommand\aucollab[1]{\unskip}
  
\@ifundefined{tsGraphicsScaleX}{\gdef\tsGraphicsScaleX{1}}{}
\@ifundefined{tsGraphicsScaleY}{\gdef\tsGraphicsScaleY{.9}}{}
\def\checkGraphicsWidth{\ifdim\Gin@nat@width>\linewidth
	\tsGraphicsScaleX\linewidth\else\Gin@nat@width\fi}

\def\checkGraphicsHeight{\ifdim\Gin@nat@height>.9\textheight
	\tsGraphicsScaleY\textheight\else\Gin@nat@height\fi}

\def\fixFloatSize#1{}%\@ifundefined{processdelayedfloats}{\setbox0=\hbox{\includegraphics{#1}}\ifnum\wd0<\columnwidth\relax\renewenvironment{figure*}{\begin{figure}}{\end{figure}}\fi}{}}
\let\ts@includegraphics\includegraphics

\def\inlinegraphic[#1]#2{{\edef\@tempa{#1}\edef\baseline@shift{\ifx\@tempa\@empty0\else#1\fi}\edef\tempZ{\the\numexpr(\numexpr(\baseline@shift*\f@size/100))}\protect\raisebox{\tempZ pt}{\ts@includegraphics{#2}}}}

\AtBeginDocument{\def\includegraphics{\@ifnextchar[{\ts@includegraphics}{\ts@includegraphics[width=\checkGraphicsWidth,height=\checkGraphicsHeight,keepaspectratio]}}}

\DeclareMathAlphabet{\mathpzc}{OT1}{pzc}{m}{it}

\def\URL#1#2{\@ifundefined{href}{#2}{\href{#1}{#2}}}

\def\UrlOrds{\do\*\do\-\do\~\do\'\do\"\do\-}%
\g@addto@macro{\UrlBreaks}{\UrlOrds}

\edef\fntEncoding{\f@encoding}

\makeatother

\newif\ifmultipleabstract\multipleabstractfalse%
\makeatletter
\AtBeginDocument{\@ifpackageloaded{longtable}{%
\def\LT@makecaption#1#2#3{%
  \LT@mcol\LT@cols c{\hbox to\z@{\hss\parbox[t]\LTcapwidth{%
    \sbox\@tempboxa{#1{#2: } #3}%
    \ifdim\wd\@tempboxa>\hsize
      #1{#2: }\textsc{#3}%
    \else
      \hbox to\hsize{\hfil\box\@tempboxa\hfil}%
    \fi
    \endgraf\vskip\baselineskip}%
  \hss}}}
}{}}
\makeatother

\makeatletter
\let\citep\cite
\let\citet\cite
\makeatother

\begin{document}


        \title{Ethical Challenges in Computer Vision: Ensuring Privacy and Mitigating Bias in Publicly Available Datasets}
      
\author{Ghalib~Ahmed~Tahir\thanks{Ghalib Ahmed~Tahir is with University Malaya,
        Kuala Lumpur,
        Kuala Lumpur,
        -,
        Federal,
        Malaysia, e-mail: ghalib@siswa.um.edu.my (Corresponding author).}}

\maketitle 

\begin{abstract}
This paper aims to shed light on the ethical problems of creating and deploying computer vision technology, particularly in using publicly available datasets. Due to the rapid growth of machine learning and artificial intelligence, computer vision has become a vital tool in many industries, including medical care, security systems, and trade. However, extensive use of visual data that is often collected without consent due to an informed discussion of its ramifications raises significant concerns about privacy and bias. The paper also examines these issues by analyzing popular datasets such as COCO, LFW, ImageNet, CelebA, PASCAL VOC, etc., that are usually used for training computer vision models. We offer a comprehensive ethical framework that addresses these challenges regarding the protection of individual rights, minimization of bias as well as openness and responsibility. We aim to encourage AI development that will take into account societal values as well as ethical standards to avoid any public harm. 
\end{abstract}
    
% \fi
\IEEEpeerreviewmaketitle

\section{Introduction}

\subsection{The Rise of Computer Vision and Its Societal Impact}The domain of computer vision has lately undergone an unparalleled surge due to the developments in machine learning, deep learning, and the availability of large-scale datasets. Consequently, machines can recognize and interpret visual data using these technologies, leading to breakthroughs in fields like health care\unskip~\cite{2449214:31663850} , driverless cars\unskip~\cite{2449214:31663849}, security\unskip~\cite{2449214:31663851}, and retail. For instance, medical conditions can be correctly diagnosed by computer vision systems from images with very high accuracy, whereas facial recognition \unskip~\cite{2449214:31663853} is used for enhancing security and guiding autonomous vehicles.

However, the more that these technologies become part of everyday life brings significant ethical issues to bear. This pervasive gathering and use of visual information sometimes without the permission or awareness of those being imaged raises concerns about privacy and possible misuse. Additionally, bias ingrained in these databases creates discriminatory results that affect marginalized groups more significantly than other social classes. These technical problems are intrinsically connected to wider societal values such as human dignity, respect, fairness \unskip~\cite{2449214:31663854}, and justice.

\subsection{Ethical Challenges in Public Datasets}To develop computer vision systems, public datasets are necessary to train and evaluate algorithms. However, the use of these datasets often comes with ethical pitfalls. Many contain images of individuals who have not provided explicit consent for their images to be used, thus raising privacy concerns that ought to be addressed \cite{2449214:31810191}. What is more, these datasets frequently reflect societal biases, which can in turn be inadvertently operationalized by the models trained on them. For instance, the lack of diversity in training datasets has resulted in facial recognition systems being less accurate for people with dark skin \cite{2449214:31810192}. To quantify the risk of privacy breach from a dataset, one could use the following equation.
\let\saveeqnno\theequation
\let\savefrac\frac
\def\dispfrac{\displaystyle\savefrac}
\begin{eqnarray}
\let\frac\dispfrac
\gdef\theequation{1}
\let\theHequation\theequation
\label{dfg-27187b9eb1b7}
\begin{array}{@{}l}PrivacyRisk\;=\;\sum_{i\;=1}^{N}P(Reidentification\mid DataPoint_i)\end{array}
\end{eqnarray}
\global\let\theequation\saveeqnno
\addtocounter{equation}{-1}\ignorespaces 
Where,

N is the total number of points in the dataset.

$P(Reidentification\mid DataPoint_i) $

is the probability that an individual can be re-identified from data point.

Building computer vision datasets often overlooks the concept of informed consent, which is a keystone of ethical research. People whose images are included might not know that they are involved or even what it would be used for or who would use it \cite{2449214:31810193}. This lack of transparency and accountability is ethically problematic, especially when such technologies are applied in areas like surveillance and law enforcement agencies \cite{2449214:31810194}.

\subsection{Objective}This paper will seek to propose a regulated manner of including ethical principles in the design and use of computer vision datasets. The focus will be on privacy, bias reduction, and transparency as a way of solving the identified ethical problems with AI guidelines for developers. This work is aimed at making sure that the growth of computer vision systems matches societal values and moral norms thereby promoting dependability and openness in using these powerful technologies.
    
\section{Ethical Principles in Computer Vision}

\subsection{Respect for Human Dignity and Privacy}The development and use of computer vision technologies must be guided by the ethical principle that respects human dignity and privacy. In publicly available datasets, this requires that the rights of individuals to privacy are observed throughout the life-cycle of data, from sample collection to model deployment \cite{2449214:31810228}. Some of these include obtaining informed consent whenever possible, applying anonymization techniques and ensuring that people are not exploited or harmed through their images \cite{2449214:31810229}.

\subsubsection{Informed Consent}For human dignity and respect, informed consent is paramount. It guarantees that people know how their images will be used and have a choice to opt-out if they want to \cite{2449214:31810230}. Nonetheless, getting informed consent can be problematic especially where large-scale datasets are concerned in which internet-scraped or public space-captured pictures may lack the awareness of those portrayed \cite{2449214:31810233}. This problem is worse when pictures go beyond their initial context e.g. commercial uses or research applications against the expectation of any individual involved \cite{2449214:31810234}.

\subsubsection{Anonymization Techniques}To protect individuals' privacy, anonymization involves the extraction or masking of personally identifiable information (PII) from datasets. For instance, this can be achieved by applying techniques like blurring faces, removing metadata, or replacing real images with synthetic data in computer vision. However, it must be noted that anonymization is not foolproof and re-identification advancements pose a great challenge to maintaining anonymity necessitating regular updating of anonymization methods as well as continuous audits on datasets \cite{2449214:31810235} \cite{2449214:31810236}. To evaluate the effectiveness of anonymization techniques, you can define the success rate of anonymity as.
\let\saveeqnno\theequation
\let\savefrac\frac
\def\dispfrac{\displaystyle\savefrac}
\begin{eqnarray}
\let\frac\dispfrac
\gdef\theequation{2}
\let\theHequation\theequation
\label{dfg-629027ca08da}
\begin{array}{@{}l}AnonymizationSuccess=1-\;\frac{Number\;of\;Reidentified\;Individuals\;\noindent }N\end{array}
\end{eqnarray}
\global\let\theequation\saveeqnno
\addtocounter{equation}{-1}\ignorespaces

\subsubsection{Transparency in Data Usage}Computer vision systems must be transparent if they are to be trusted \cite{2449214:31810238}. The developers should also state openly how data is collected, processed, and used. It also involves providing extensive documentation about ethical considerations taken when creating datasets and measures put in place to ensure privacy together with potential risks that may arise out of the use of such data \cite{2449214:31810193}.

\subsection{Bias Prevention and Fairness}The development of computer vision models necessitates prevention of the bias and consideration for fairness. In the development of these models, various ways can introduce bias into datasets such as; overrepresentation or underrepresentation of certain communities, biased labeling practices, or selection of biased training data \cite{2449214:31810239}. The use of biased datasets to train models may mean that the systems created will maintain or worsen societal gaps \cite{2449214:31810192} \cite{2449214:31810240}. To measure bias in classifier, one could use the Disparate Impact ratio.
\let\saveeqnno\theequation
\let\savefrac\frac
\def\dispfrac{\displaystyle\savefrac}
\begin{eqnarray}
\let\frac\dispfrac
\gdef\theequation{3}
\let\theHequation\theequation
\label{dfg-f422a219a730}
\begin{array}{@{}l}DI\;=\;\frac{P(Y\;=1\mid A=1)}{P(Y\;=1\mid A=0)}\end{array}
\end{eqnarray}
\global\let\theequation\saveeqnno
\addtocounter{equation}{-1}\ignorespaces 
Where:

Y is the predicted outcome. A is a binary sensitive attribute (e.g., gender, race). P(Y=1\ensuremath{\mid }A=1) is the probability of a positive outcome for the advantaged group. P(Y=1\ensuremath{\mid }A=0) is the probability of a positive outcome for the disadvantaged group.

\subsubsection{What causes bias?}Bias in computer vision datasets can come from different sources. For instance, a dataset that mostly includes photos of light-skinned people might result in facial recognition models that are bad at recognizing individuals with dark skin. In addition, some traits are frequently associated with particular demographic groups because it is evident from uneven labeling practices that support stereotypes considered unhealthy \cite{2449214:31810241}.

\subsubsection{Bias Identification and Alleviation}It is important to ensure fairness and equality while detecting and mitigating bias in the computer vision models. This involves examining datasets for potential biases, implementing strategies that fix identified biases, and evaluating model fairness before deploying them \cite{2449214:31810242}. Among these techniques include re-sampling, re-weighting, and adversarial debiasing which serve as tools to address data set and model biases \cite{2449214:31810243}.

\subsubsection{Fairness Metrics}For instance, fairness metrics that researchers employ to evaluate the fairness of computer vision models are Disparate Impact, Demographic Parity, and Equal Opportunity \cite{2449214:31810245}. They should make use of these metrics as a way of quantifying the level of equity in terms of how different demographic groups are treated by the model so that they can be used to create fairer models \cite{2449214:31810240}.

\subsubsection{Ethical Implications}The ethical implications of bias in computer vision models are enormous. Some biased models may result in such discriminatory practices as misidentification in facial recognition systems that would have serious effects on individuals and communities \cite{2449214:31810246}. This means that addressing bias is not just a technical problem but also an ethical duty requiring a multi-disciplinary endeavor incorporating social sciences, ethics, and legal studies \cite{2449214:31810247}.

\section{Case Studies of Publicly Available Datasets: Ethical Considerations}
This section provides an in-depth analysis of several widely used computer vision datasets, highlighting the ethical considerations associated with each. We focus on privacy, bias, and the broader societal implications of using these datasets in AI development.

\subsection{COCO (Common Objects in Context)}COCO\unskip~\cite{2449214:31653997}  is a vast object detection, segmentation, and captioning database that encompasses over three hundred and thirty thousand pictures with more than two and a half million labeled objects belonging to eighty types. The images are of everyday scenes and things from the internet.

For instance, COCO images are obtained without the consent of the subject \cite{2449214:31810249}, raising serious privacy concerns. Most of these images have identifiable faces and other distinctive features which poses ethical questions on whether these images can be used to train AI models. Furthermore, although it may present as diverse in terms of objects, this diversity might hide underlying biases in its representation. It may be inappropriate for example to find out that some objects disproportionately correspond to particular cultural or demographic contexts leading to biased model predictions.

To address these ethical issues, our recommendations are implementing anonymization techniques for privacy protection with individuals. These include using tools for face or other identifying marks detection and blurring them on photographs. Another approach will involve applying bias-detecting tools in order to ensure that the dataset encompasses all demographics equally. Ethical compliance within the dataset should be audited regularly through conducting checks and updates

% \unskip~\cite{2449214:31653999}

\subsection{LFW (Labeled Faces in the Wild) What is LFW?}LFW [28]  is a face recognition benchmark dataset that has over 13,000 labeled facial images. It is widely used in studies on face recognition, and it's considered particularly difficult because the pictures have been taken in uncontrolled settings.

\subsubsection{Ethical Implications}Most of these faces are identifiable, which raises substantial privacy concerns especially given that very many people who are listed there have not provided explicit consent for their photos to be used. Additionally, LFW mainly features public figures, which can introduce bias in facial recognition models if the dataset does not represent the general population well.

\subsubsection{Recommendations}We recommend employing robust techniques for anonymization, such as blurring of faces and revisiting the demography of the dataset. Besides, it should also be considered to exclude pictures where consent is questionable. Thus, mitigation and detection measures regarding bias are required to ensure that LFW-trained models are fair and inclusive.

\subsection{ImageNet}ImageNet \unskip~\cite{2449214:31663847} is among the most popular datasets for computer vision problems, which consists of over 14 million images and is categorized into more than 20,000 classes. These tasks have contributed massively to image-related classification efforts while being applied in training some of the most successful deep-learning models.

\paragraph{Ethical Issues}The dataset's labels have faced criticisms for their bias and inclusion of irrelevant categories. However, the extent to which each category can be reviewed has hindered a thorough review of its entire list; thus, it may have retained certain terms reinforcing harmful attitudes or biased stereotypes.

\subsubsection{Recommendations}ImageNet should be audited regularly and updated so that biased or inappropriate content is removed \cite{2449214:31810251}. A systematic ethical examination process should be established to continually evaluate whether the dataset complies with moral norms. Also, bias identification and reduction tools must accompany its creation and use so as not to perpetuate social inequalities \cite{2449214:31810252}.

% \unskip~\cite{2449214:31658009}

\subsection{CelebA (CelebFaces Attributes Dataset)}CelebA's overview: Over 200,000 celebrity images [32] with 40 attributes, including gender, age, and some facial details have been labeled. It is mostly used in the task of facial attribute recognition as well as research on generative models and face manipulation methods.

\subsubsection{Ethical issues}The problem of privacy is presented when these images are used for commercial purposes without the explicit permission from celebrities who own them. The attribute labels in CelebA can be reinforcing stereotypes, particularly if trained models on the dataset are employed to predict attributes in non-celebrity images.

\subsubsection{Recommendations}It would be appropriate to re-evaluate the ethical implications of using such photographs, especially in commercial works. Bias mitigation techniques should be put in place to prevent these models trained with CelebA from perpetuating harmful stereotypes. Transparency in its application documentation is paramount for maintaining strong ethics.

\subsection{PASCAL VOC (Visual Object Classes)}

\subsubsection{Review}A standard dataset for visual object classification and detection, PASCAL VOC \unskip~\cite{2449214:31663845} contains pictures of 20 categories of objects. It has been extensively adopted in the field of computer vision studies as a benchmark for object detection models.

\subsubsection{Moral Dilemmas}Informed consent and potential bias in object representation are key concerns that affect similar datasets. There are numerous sources of images in PASCAL VOC, sometimes making it hard to tell whether the depicted individuals or entities gave their permission to use their images.

\subsubsection{Suggestions}These ethical challenges can be addressed by implementing privacy-preserving techniques and regularly evaluating the dataset for any biases. For responsible deployment purposes, ensuring that PASCAL VOC fairly represents a wide range of objects and contexts is important in coming up with impartial models.
    
\section{Proposed Ethical Framework for Computer Vision Dataset Development}
Our proposed framework should address this issue. This will serve as a guideline for researchers, developers and policy makers to ensure computer vision is founded on ethical principles and social values.

\subsection{Informed Consent and Anonymization}

\subsubsection{Guideline}Wherever possible, consent should be sought from individuals whose images are captured in datasets. In case it is not possible, techniques of anonymization should be applied so as to conceal identities.

\subsubsection{Rationale}Overall, informed consent respects the autonomy of human beings while at the same time upholding their dignity. It provides an opportunity for subjects to decide on usage or opt out when they want to protect their image's integrity. When there's an absence of informed consent, anonymization allows for further use of the dataset without compromising privacy interests thus serving as a secondary protection \cite{2449214:31810251}.

\subsubsection{Tool Implementation}The recommendation is therefore to include automated tools that can identify and then hide any identifiable parts within the images (Nayak et al., 2018). For example, facial blurring can be used to blur faces in photos thereby reducing chances of identity theft through photographs being posted online. Moreover, metadata linked with these pictures has to be either stripped off or disguised, hence is eliminate revealing any form of personal information related to them.

\subsubsection{Challenges and Considerations}The completeness of anonymization does not guarantee its effectiveness and it is possible that one can link the anonymous data to specific individuals through sophisticated re-identification techniques. Therefore, it becomes important that we keep updating our anonymization methods while keeping a check on whether our datasets adhere to privacy standards or not.

\subsection{Bias Detection and Mitigation}

\subsubsection{Guideline}Regularly evaluate data sets for gender, race, age, or other protected characteristics-related biases. Implement ways of mitigating these biases, including through resampling or reweighting the datasets.

\subsubsection{Rationale} The presence of bias within datasets can result in models that are unfair across different demographic categories, thereby maintaining or exacerbating existing social inequalities. Detecting and diminishing bias is a crucial requirement for creating fair and just machine learning models on computer vision that will benefit all communities.

\subsubsection{Tool Implementation}The dataset development process should include modules for detecting biases. These may be used to assess demographic diversity within the datasets. Other methods like undersampling, redistribution of weights, and synthetic data generation could level the playing field by removing imbalances among various racial populations.
\let\saveeqnno\theequation
\let\savefrac\frac
\def\dispfrac{\displaystyle\savefrac}
\begin{eqnarray}
\let\frac\dispfrac
\gdef\theequation{4}
\let\theHequation\theequation
\label{dfg-52bb0bbf920f}
\begin{array}{@{}l}w_i\;=\;\frac1{P(A_i)}\end{array}
\end{eqnarray}
\global\let\theequation\saveeqnno
\addtocounter{equation}{-1}\ignorespaces 
Where,

$w_i $ \noindent  is the weight assigned to sample i. $P(A_i) $ is the probability of the sensitive attribute A in the original dataset.

\subsubsection{Challenges and Considerations}Bias detection, as well as mitigation, is not an easy undertaking since it requires in-depth knowledge about what causes the bias itself. There could be different strategies for countering various types of bias. Seek to avoid perpetuating new discriminatory practices or reducing model performance during attempts to correct previous ones.

\subsection{Content Filtering That Is Sensitive}

\subsubsection{Guideline}Leave out or mark sensitive content that may be unsafe or inappropriate for data suitability consideration.

\subsubsection{Rationale}Inappropriate, sensitive content in datasets can harm people and communities. It is important to filter such content as it protects people while responsibly utilizing the datasets.

\subsubsection{Tool Implementation}For instance, one can develop and apply filter algorithms that are specific to an identified dataset. Some algorithms can be trained to identify violent, pornographic, or symbols of hate like nudity in images and flag them accordingly. Additionally, flagged content must be reviewed manually \cite{2449214:31810254}.

\subsubsection{Challenges and Considerations}One of the challenges with implementing content filtering on large-scale datasets, is cost-effectiveness. On other occasions, manual review processes may take longer since automated algorithms might inaccurately tag a sensitive part of information. Therefore, there should always be continuous refinement of filtration algorithms and standards on how flagged materials should be dealt with \cite{2449214:31810255}.

F be the fraction of the dataset flagged by the automated algorithm as sensitive. M be the manual review rate, i.e., the fraction of flagged data that can be manually reviewed per unit time. \mbox{}\protect\newline The time  needed for filtering and reviewing the entire dataset can be expressed as:
\let\saveeqnno\theequation
\let\savefrac\frac
\def\dispfrac{\displaystyle\savefrac}
\begin{eqnarray}
\let\frac\dispfrac
\gdef\theequation{5}
\let\theHequation\theequation
\label{dfg-65275e4175c4}
\begin{array}{@{}l}T_f\;=\;\frac{D\ast F}M\end{array}
\end{eqnarray}
\global\let\theequation\saveeqnno
\addtocounter{equation}{-1}\ignorespaces

\subsection{Transparence And Documentation}

\subsubsection{Guideline}Clearly indicate where images are obtained from, how they were collected curated, or cleaned up, and ethical matters are taken into account.

\subsubsection{Rationale}Transparency is important for creating trust in computer vision systems and using datasets with fairness. Transparency must be maintained to achieve this.

\subsubsection{Tool Implementation}There should be an ethical implementation report template that guides the documentation process during dataset development. It will provide details about data sources, consent mechanisms, anonymity mechanisms adopted, how to detect bias and mitigate it as well as filtering of its content. Finally, it has to be open to public scrutiny thus increasing accountability.

\subsubsection{Challenges and Considerations}Exhaustive documentation can be difficult especially when dealing with large datasets, which have complex processes of collecting and curating information. This implies that adequate resources ought to be allocated towards, the documentation efforts while ensuring that they are updated continuously with emerging ethical issues.

\subsection{Regular Audits and Updates}

\subsubsection{Guideline} Ethical guidelines conformity should be audited periodically using datasets and new ethical challenges should be addressed.

\subsubsection{Rationale}Computer vision faces contemporary ethical problems that necessitate constant auditing to ensure the alignment of datasets with moral principles. This way, we can be able to detect possible issues in advance and avoid them from growing into huge challenges.

\subsubsection{Tool Implementation}It is essential that an automated auditing tool be implemented to check if datasets comply with ethical codes. Datasets must be checked for privacy threats, prejudices towards specific groups or individuals, as well as objectionable content, and then a report will be provided on its findings. Moreover, the tool must suggest necessary alterations in order to remove any issues identified before.

\subsubsection{Challenges and Considerations} Ongoing dedication is needed for regular audits and therefore integrating auditing procedures into the overall dataset-building process is quite important. Besides that, audits related to ethics as well as computer vision should never be done by amateurs; instead, such processes call for experts who are competent enough in this area.

To quantify the ethical compliance of a dataset:
\let\saveeqnno\theequation
\let\savefrac\frac
\def\dispfrac{\displaystyle\savefrac}
\begin{eqnarray}
\let\frac\dispfrac
\gdef\theequation{6}
\let\theHequation\theequation
\label{dfg-63b4608c8c7c}
\begin{array}{@{}l}ComplianceScore=\frac1M\sum_{j\;=\;1}^{M}ComplianceMetric_j\end{array}
\end{eqnarray}
\global\let\theequation\saveeqnno
\addtocounter{equation}{-1}\ignorespaces 
Where,

M is the number of ethical metrics considered (e.g., privacy, bias, transparency).  Compliance Metric j is a normalized score (0 to 1) indicating how well the dataset meets the jth ethical criterion.
    
\section{Conclusion}

\subsection{Summary}The study aims to create a comprehensive framework for incorporating ethical principles into computer vision dataset development and usage. We do this by emphasizing human dignity and privacy, bias mitigation, and transparency that offers practical guidelines and tools to AI developers in order to make datasets corresponding with ethical requirements. Our framework is intended to overcome ethical dilemmas faced due to the utilization of publicly available datasets in computer vision, to ensure these powerful technologies are responsibly developed and deployed.

\subsection{Future Work}Further studies should seek better techniques to identify and reduce discrimination while involving new ethical issues that rise with the advancement of computer vision technology. Additionally, the applicability and efficiency of proposed framework can be tested on other types of datasets in different applications.

\subsection{Possible ramifications for the development of AI}AI developers can guarantee that, besides being technically sound, this will also ensure that their models are ethically responsible if they embrace these ethical practices. This ensures trust in AI technologies and also encourages their ethical application in society. The ever-increasing computer vision necessitates that ethical issues be the cornerstone during dataset creation and model deployment. Meeting such obstacles is not only a technical necessity but also a moral obligation which shows our shared responsibility towards building technological systems that foster societal benefit.



%\bibliographystyle{IEEEtran}

\bibliographystyle{IEEEtran}

\bibliography{article}
\vfill
\end{document}